\documentclass{article}

\usepackage{graphicx}
\usepackage{subcaption}
\usepackage{microtype}
\usepackage{booktabs} %
\usepackage{float}
\usepackage{hyperref}

\usepackage[accepted]{icml2026}

\usepackage{amsmath}
\usepackage{amssymb}
\usepackage{mathtools}
\usepackage{amsthm}

\usepackage[capitalize,noabbrev]{cleveref}

\theoremstyle{plain}

\theoremstyle{definition}

\theoremstyle{remark}

\usepackage[textsize=tiny]{todonotes}

\icmltitlerunning{Emergent Compositional Skills in VLAs}

\begin{document}

\twocolumn[
  \icmltitle{Emergent Compositional Skills in Mixture-of-Experts VLAs}

  \icmlsetsymbol{equal}{*}

  \begin{icmlauthorlist}
    \icmlauthor{ Shlok Shah}{equal,cs}
    \icmlauthor{ Rhiaan Jhaveri}{equal,cs}
    \icmlauthor{ Tharun Kumar Tiruppali Kalidoss}{equal,cs}
    \icmlauthor{ Chirayu Nimonkar}{equal,cs}
    \icmlauthor{ Ishaan Javali}{equal,cs}
    \icmlauthor{ Dhruv Shah}{ece}
  \end{icmlauthorlist}

  \icmlaffiliation{cs}{Department of Computer Science, Princeton University, Princeton, NJ, USA}
  \icmlaffiliation{ece}{Department of Electrical and Computer Engineering, Princeton University, Princeton, NJ, USA}

  \icmlcorrespondingauthor{Tharun Kumar Tiruppali Kalidoss}{tharun.tiruppali@gmail.com}

  \icmlkeywords{Vision-Language-Action Models, Mixture-of-Experts, Compositional Learning}

  \vskip 0.3in
]

\makeatletter
\gdef\@icmltitlerunning{Emergent Compositional Skills in Mixture-of-Experts VLAs}
\makeatother

\printAffiliationsAndNotice{\icmlEqualContribution}

\begin{abstract}

We consider the problem of learning compositional robot policies end-to-end from expert demonstrations, without any pre-specified notion of task decomposition or hierarchy. We ask whether a VLA trained with a simplified Mixture-of-Experts (MoE) 
action head can emergently learn to decompose tasks into reusable, interpretable primitives. We find that learned experts are heavily reused across tasks and consistently correspond to 
qualitatively distinct low-level behaviors, suggesting that the router 
implicitly learns to perform high-level sequencing
while experts serve as 
compositional primitives. Our MoE matches the task performance of a 
monolithic baseline while demonstrating meaningful expert specialization, a step toward modular, interpretable robot policies that emerge from 
data alone.

\end{abstract}

\section{Introduction}

Vision-language-action models (VLAs) have shown promise for generalizing across a wide range of robotic manipulation tasks \cite{rt2, openx, octo, openvla}. However, VLAs are typically trained and deployed as monolithic policies, making it difficult to identify reusable skills, compose behaviors hierarchically, or adapt parts of the policy without modifying the entire model. In contrast, many robotic tasks naturally decompose into reusable behavioral modes, such as reaching, grasping, and placing. A modular policy that explicitly learns such skills could improve interpretability, compositionality, and robustness while retaining the broad task competence of VLAs. We study whether this modular structure can be learned directly from data.

Multiple expert policies are trained end-to-end with a learned router that selects among experts conditioned on the current observation and language instruction. The router acts as an implicit high-level controller, while the experts specialize into lower-level behavioral modes, forming an emergent hierarchy without requiring a pre-specified decomposition or manually defined skill library. This contrasts with prior hierarchical VLA systems that impose a fixed planner-controller split \cite{hamster, fast-slow, rlt}, and offers potential benefits for long-horizon planning and transfer to new tasks.

We find that even a simple Mixture-of-Experts action expert architecture 
produces experts that learn distinguishable, reusable primitives while maintaining comparable performance against finetuning the baseline.

\vspace{-0.5em}
\section{Approach}
\label{sec:approach}

\begin{figure*}[t]
    \centering
    \includegraphics[width=\linewidth]{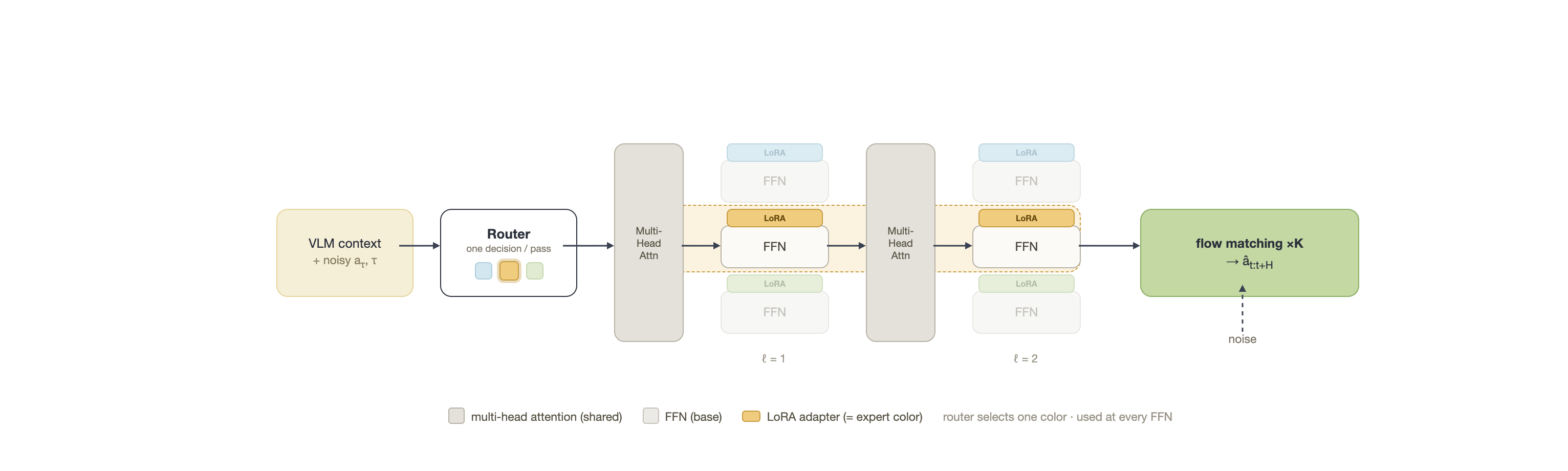}
    \caption{\textbf{LoRA Mixture-of-Experts architecture.} The router makes a single decision per forward pass from the VLM context (plus the noised action chunk $a_\tau$ and flow-matching timestep $\tau$), selecting one expert that is applied at every action-expert layer. Multi-head attention is shared across experts; only the FFN sublayer is replaced by a base FFN plus a routed LoRA adapter. The same expert (color) is used at every FFN, and the action chunk $\hat{a}_{t:t+H}$ is produced through $K$ flow-matching steps.}
    \label{fig:architecture}
\end{figure*}
\label{sec:approach}

\label{expert_skills}
\begin{figure*}[t]
    \centering
    \begin{subfigure}[t]{0.32\textwidth}
        \centering
        \includegraphics[width=0.85\linewidth]{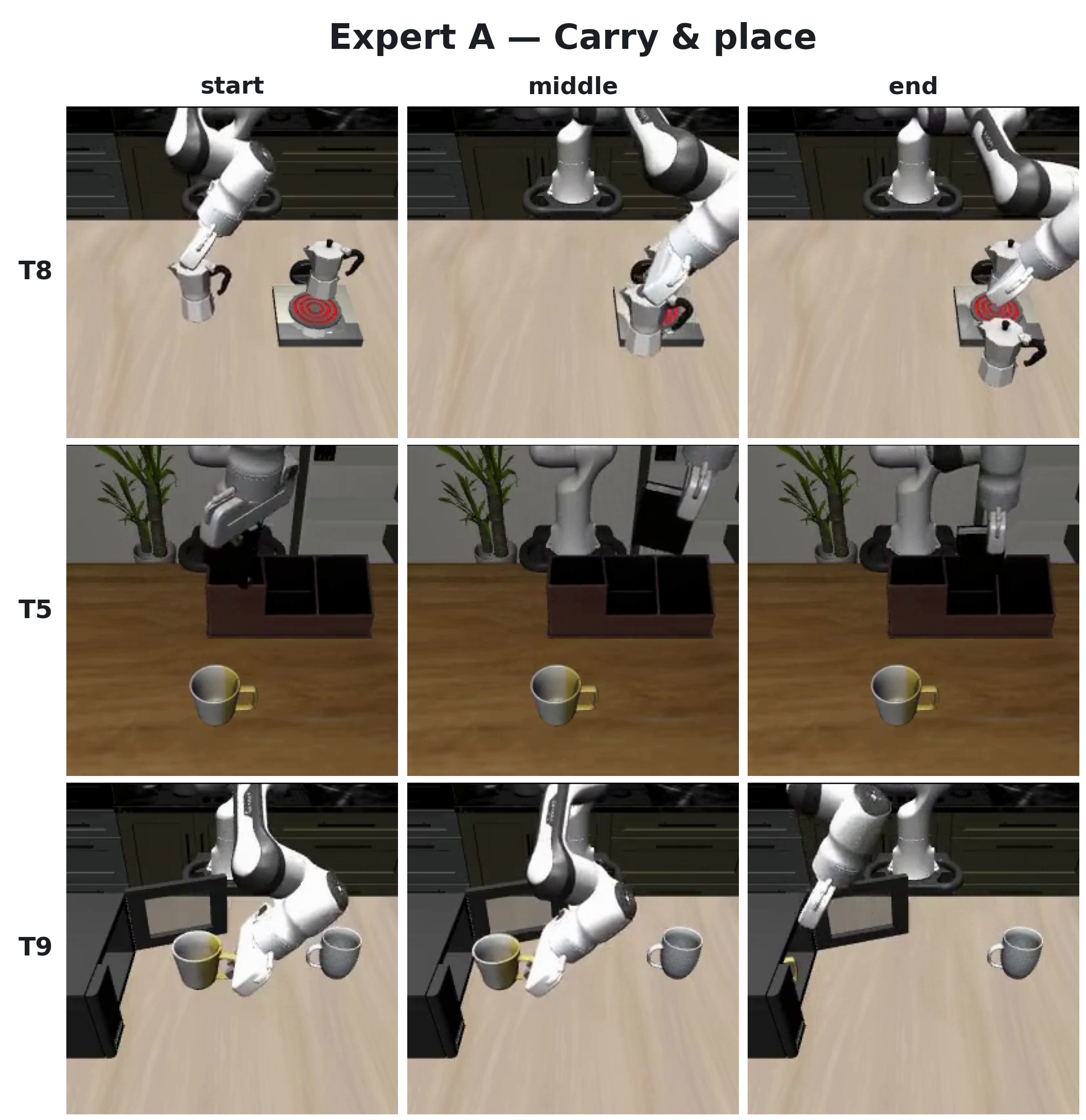}
        \caption{Expert A \textbf{places a grasped object at its final position}.}
        \label{fig:expert_a}
    \end{subfigure}
    \hfill
    \begin{subfigure}[t]{0.32\textwidth}
        \centering
        \includegraphics[width=0.85\linewidth]{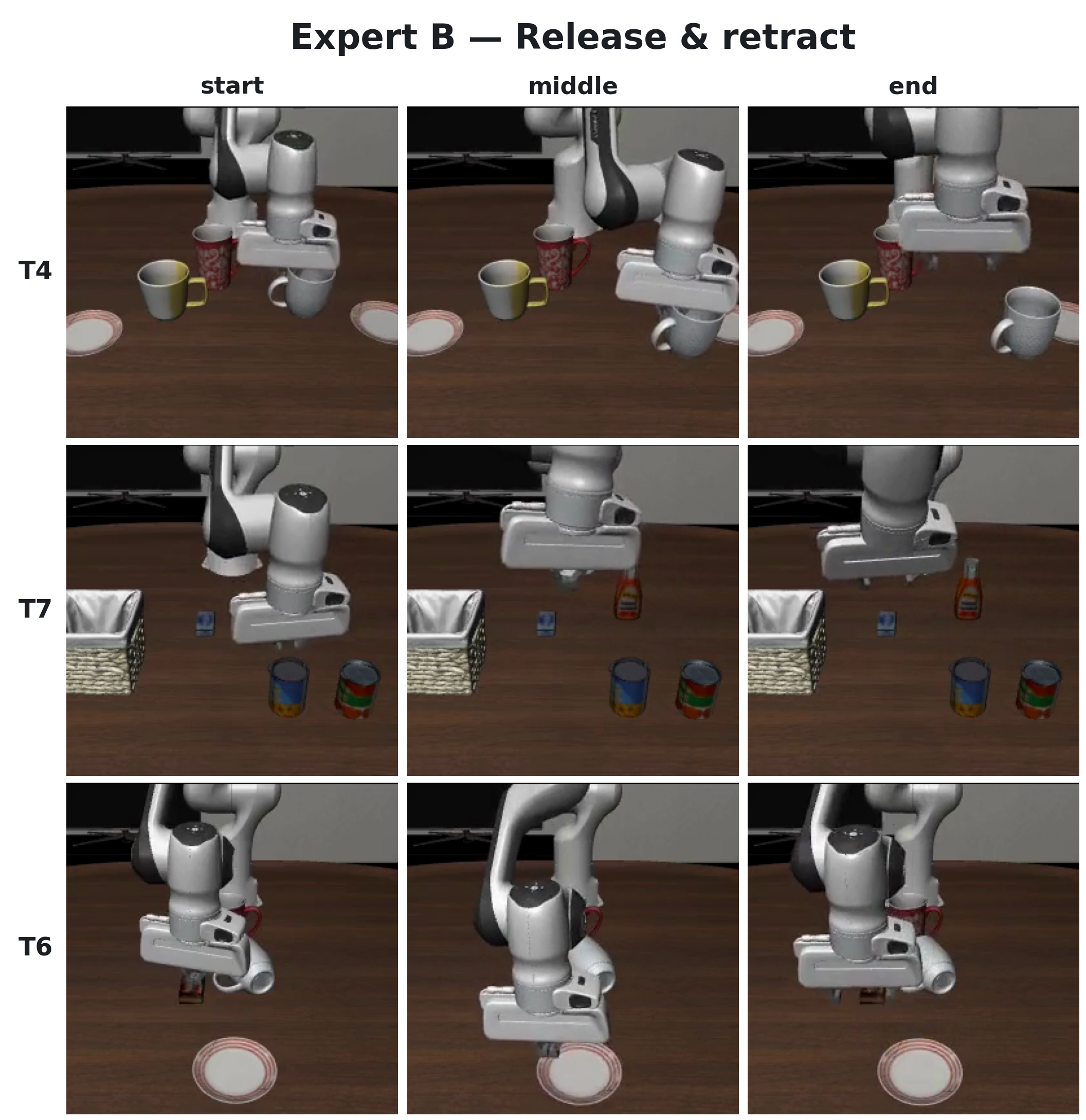}
        \caption{Expert B \textbf{releases an item and retracts upward}.}
        \label{fig:expert_b}
    \end{subfigure}
    \hfill
    \begin{subfigure}[t]{0.32\textwidth}
        \centering
        \includegraphics[width=0.85\linewidth]{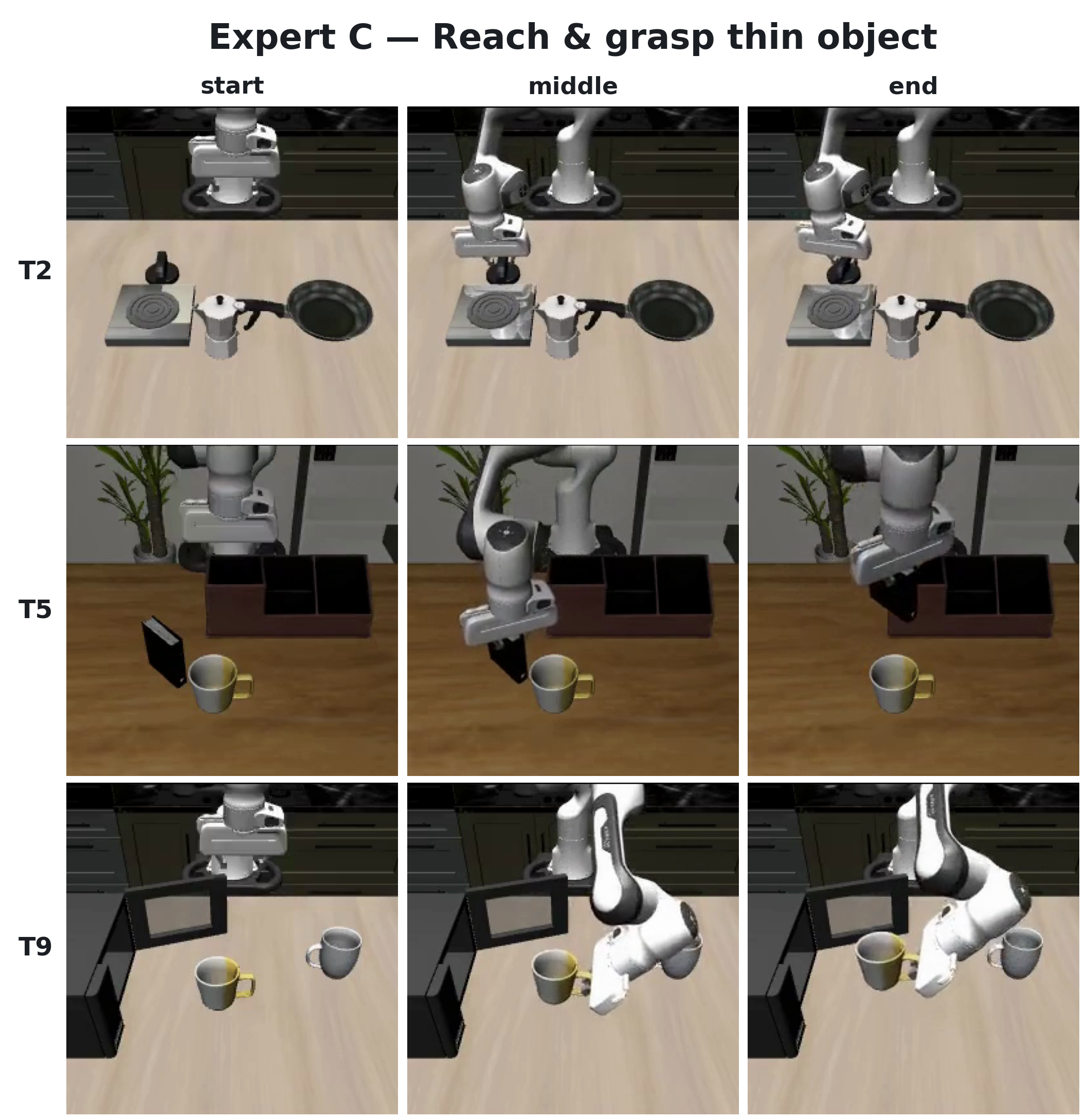}
        \caption{Expert C \textbf{approaches and grasps thin handled items}.}
        \label{fig:expert_c}
    \end{subfigure}
    \caption{\textbf{Qualitative comparison of three learned experts on LIBERO-10 rollouts.} Each column shows start, middle, and end frames of trajectories where the corresponding expert was selected by the router.}
    \label{fig:expert_skills}
\end{figure*}

We train a Mixture-of-Experts (MoE) action policy on top of two pretrained
VLA backbones, $\pi_{0}$~\citep{pi0} and SmolVLA~\citep{smolvla}, both
of which use flow matching for action generation.
Two design choices define our approach: (i) experts are realized as
low-rank (LoRA) deltas ($r = 16$) on the action expert's FFN
sublayers, giving a strong shared prior; and (ii) routing is performed
\emph{once per forward pass} from the joint visual-language-state context, so a routing decision selects a coherent skill executed by a single action expert.

\subsection{LoRA Mixture-of-Experts}
\label{sec:approach:backbone}

$\pi_{0}$~\citep{pi0} and
SmolVLA~\citep{smolvla}, each consist of a vision-language module that
encodes camera views and language into context tokens, and an action expert
(transformer decoder) that processes a proprioceptive state token and a
noised action chunk. Our method is backbone-agnostic and we apply it to
both models in our experiments (Sec.~\ref{sec:experiments}).

The MoE replaces \emph{only} the FFN sublayer of each action layer;
self-attention is shared across all experts:
\begin{equation}
\begin{split}
\mathrm{FFN}_\ell(x) = \mathrm{FFN}^{\text{base}}_\ell(x)
+ \sum_{e \in \mathcal{R}} w_e \bigl(\mathrm{FFN}^{(e)}_\ell(x) \\
- \mathrm{FFN}^{\text{base}}_\ell(x)\bigr),
\end{split}
\label{eq:ffn}
\end{equation}
where $\mathcal{R}$ is the routed expert set with renormalized weights
$w_e$ (only top-$k$ selected action experts have non-zero $w_e$), and each expert differs from the base only through rank-$r$ LoRA
deltas $\Delta W = (\alpha/r)\,BA$. Zero-initialized $B$ matrices ensure
the policy reproduces the pretrained backbone exactly at step zero, so
specialization emerges \emph{around} a strong prior rather than from
scratch.

\subsection{Whole forward pass routing}
\label{sec:approach:routing}

Routing is performed once per policy forward pass and shared across all
$L$ action-expert layers, in contrast to standard MoE transformers that
route per token per layer. Sharing a single $(\text{indices},
\text{weights})$ pair across the full depth makes each expert a coherent
end-to-end behavior (a ``skill'') rather than $L$ independent layer-wise
routing decisions.

A small MLP router $g_\phi$ takes as input a context vector encoding the
agent's visual, linguistic, and proprioceptive state:
\begin{equation}
c \;=\; \bigl[\,\mathrm{MeanPool}\bigl(\mathrm{VLM}(I, \ell, s)\bigr)\;\Vert\;W_s\, s\,\bigr],
\label{eq:ctx}
\end{equation}
where $I$, $\ell$, $s$ are camera observations, instruction, and
proprioceptive state, and $W_s$ is shared with the action expert. The
router emits logits over $E$ experts, from which we take top-$k$ and
renormalize their softmax weights~\cite{mixtral}; this selection is
applied uniformly across every MoE FFN. At inference, the router fires
once per action-chunk.

\subsection{Training objective}
\label{sec:approach:loss}

The full objective combines the backbone's flow-matching behavior-cloning
loss with two auxiliary terms,
\begin{equation}
\mathcal{L} \;=\; \mathcal{L}_{\text{FM}}
\;+\; \lambda_{\text{LB}}\, \mathcal{L}_{\text{LB}},
\label{eq:loss}
\end{equation}
where $\mathcal{L}_{\text{FM}}$ is the backbone's flow-matching 
velocity-prediction loss on the noised action chunk and 
$\mathcal{L}_{\text{LB}}$ is the standard load-balancing 
term~\citep{switch} that discourages routing collapse. We use 
$\lambda_{\text{LB}} = 0.01$ throughout.

\section{Experimental Results}
\label{sec:experiments}

We aim to answer four questions: \textbf{(Q1)} Does a VLA with an MoE 
action head learn experts that correspond to qualitatively distinct skills 
in a self-supervised manner? \textbf{(Q2)} Are learned skills reused 
consistently across and within tasks to perform similar behaviors? 
\textbf{(Q3)} 
How are low-level primitives composed in order to solve longer-horizon tasks?
\textbf{(Q4)} Do experts demonstrate distinct behaviors solely due to router placement or do experts cause these behaviors (especially in new contexts)?  We focus on  $\pi_{0}$ (with additional testing on SmolVLA) with our MoE architecture below.

\subsection{Qualitative Analysis of Expert Skills}

\label{sec:expert_skills}
\begin{figure*}[t]
\centering
\includegraphics[width=\linewidth]{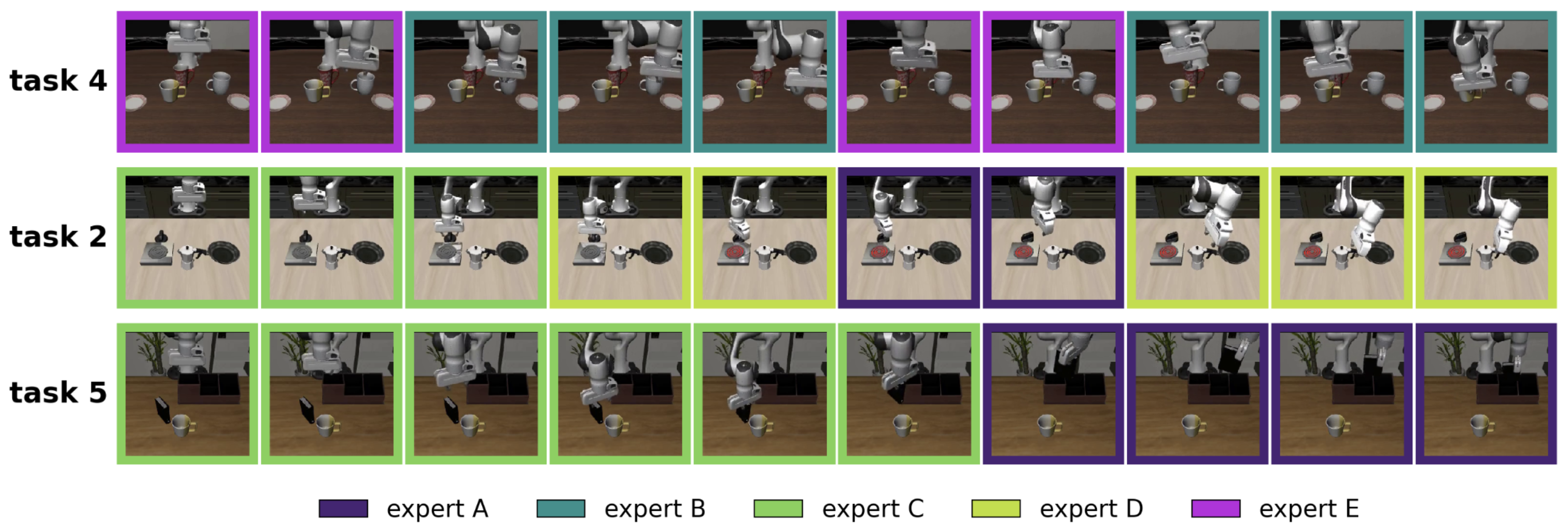}
        \caption{\textbf{Example trajectories labeled by top expert used at each step.} Different skills are composed in order to perform a more complicated, longer horizon task. In task 4,  repetition of skills allows for  recovery-like behavior.}
        \label{fig:trajectories}
\end{figure*}

Figure~\ref{fig:expert_skills} highlights three representative experts that the router selects on LIBERO-10. Expert A (Fig.~\ref{fig:expert_a}) handles the final transport phase of a manipulation, activating once an object is already grasped and carrying it to its target across three otherwise unrelated tasks (placing a moka pot on the stove, dropping a mug into a caddy, putting a yellow mug in a microwave). Expert B (Fig.~\ref{fig:expert_b}) captures the release-and-retract phase, firing after a drop-off to lift the gripper to a raised pose. Expert C (Fig.~\ref{fig:expert_c}) is selected during approach-to-grasp on objects with thin handles, such as moka pot handles and mug handles.

We observe two interesting properties of these experts. First, the same expert is reused across very different tasks and scenes (e.g., Expert A appears on stove, drawer, and microwave tasks), suggesting that the experts encode phase-level skills rather than task-specific policies. Second, the three experts shown together cover a natural manipulation cycle of approach, transport, and release, which the router stitches into a full trajectory by switching experts across denoising steps. We view these as direct qualitative evidence that 
the mixture of experts architecture
induces discrete, reusable skills rather than redundant copies of the base policy.
The load-balancing term prevents collapse onto a single expert, and the model autonomously discards capacity beyond what the task distribution actually demands (6/16 experts are laregely unused).

\subsection{Reusable vs. Task-Dependent Skills}
\begin{figure}[]
    \centering
    \includegraphics[width=\linewidth]{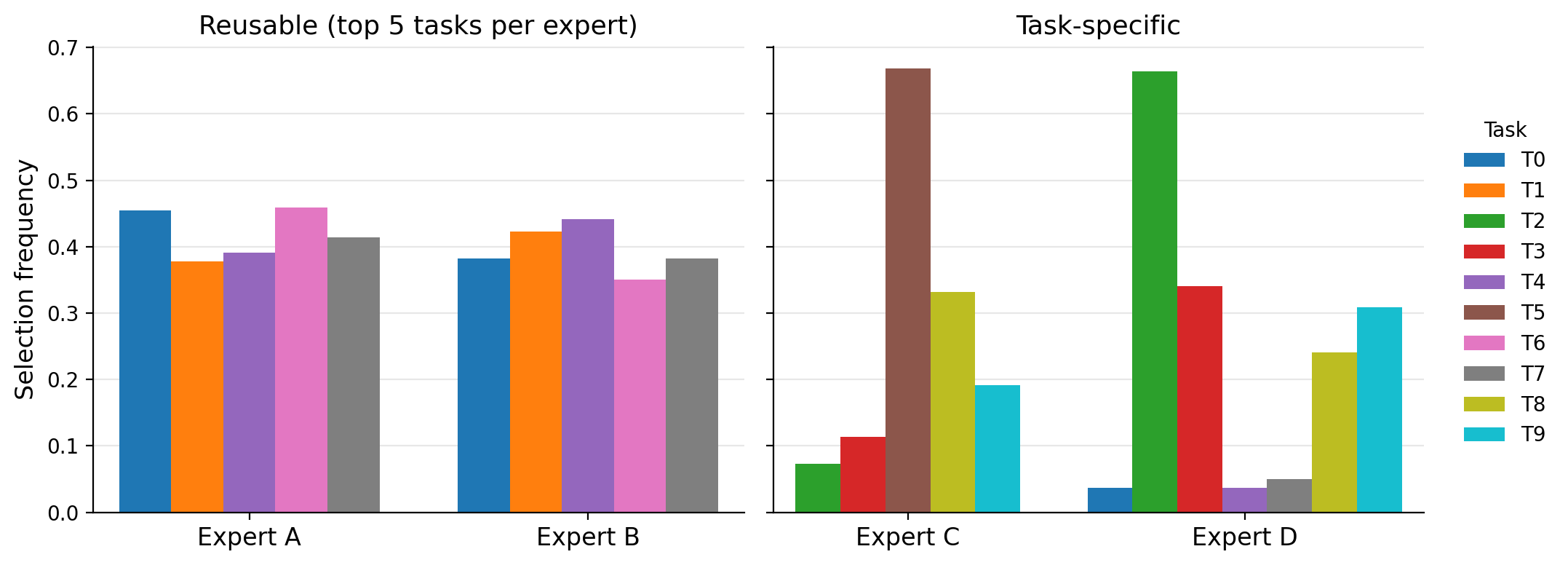}
\caption{\textbf{Selection frequency of four representative experts across LIBERO-10 tasks.} Experts C and D are \textbf{task-specific}, concentrated on T5 and T2 respectively. Bar color indicates task.}    \label{fig:expert_reuse}
    \vspace{-15pt}
\end{figure}

Inspecting how often each active expert is selected across LIBERO-10 reveals two clearly different roles. Figure~\ref{fig:expert_reuse} contrasts a representative pair of each kind.

\textbf{Reusable experts:} Experts A and B (left panel) each fire at $35\text{--}45\%$ on five different tasks and at near-zero frequency on the rest. The two cover almost the same task set (T0, T1, T4, T6, T7), and inspection of individual rollouts shows the router alternates between them within a single trajectory. This is the signature of a \emph{phase-level} skill that recurs across structurally similar tasks, matching the qualitative interpretation of the experts in Sec.~\ref{sec:expert_skills}.

\textbf{Task-specific experts:} Experts C and D (right panel) tell a different story. Expert C accounts for $67\%$ of selections on T5 but is essentially silent on the rest of the benchmark, and Expert D is similarly concentrated on T2 ($66\%$). These experts plausibly absorb idiosyncratic behaviors that only one or two tasks demand, which keeps the reusable experts crisp by sparing them the long tail of edge cases.

\label{sec:reusable}

\subsection{How are primitives composed to solve longer tasks?}

Figure~\ref{fig:trajectories} shows rollouts of the trained MoE policy on LIBERO-10 tasks T2, T4, and T5, which exhibit some of the clearest compositions of primitive skills into sequences that accomplish longer-horizon goals.

For instance, on task 5, we see that first expert C is used to grasp the book after which expert A is used to place it in its final position. On task 2, we see that expert C is used to grasp the stove dial and expert A is used to place the arm onto the handle.

Task T4 illustrates a different phenomenon: the policy fails to complete the task, but repeatedly invokes the same pair of primitives (E and C) to navigate toward the cup, attempt a grasp, move it, and release the gripper, sequencing these primitives more than five times across the trajectory. The compositional structure thus makes failures interpretable as repeated misapplication of known primitives rather than arbitrary degenerate behavior.

\subsection{Manual Routing Experiments}

\begin{figure*}[t]
    \centering
    \includegraphics[width=\linewidth]{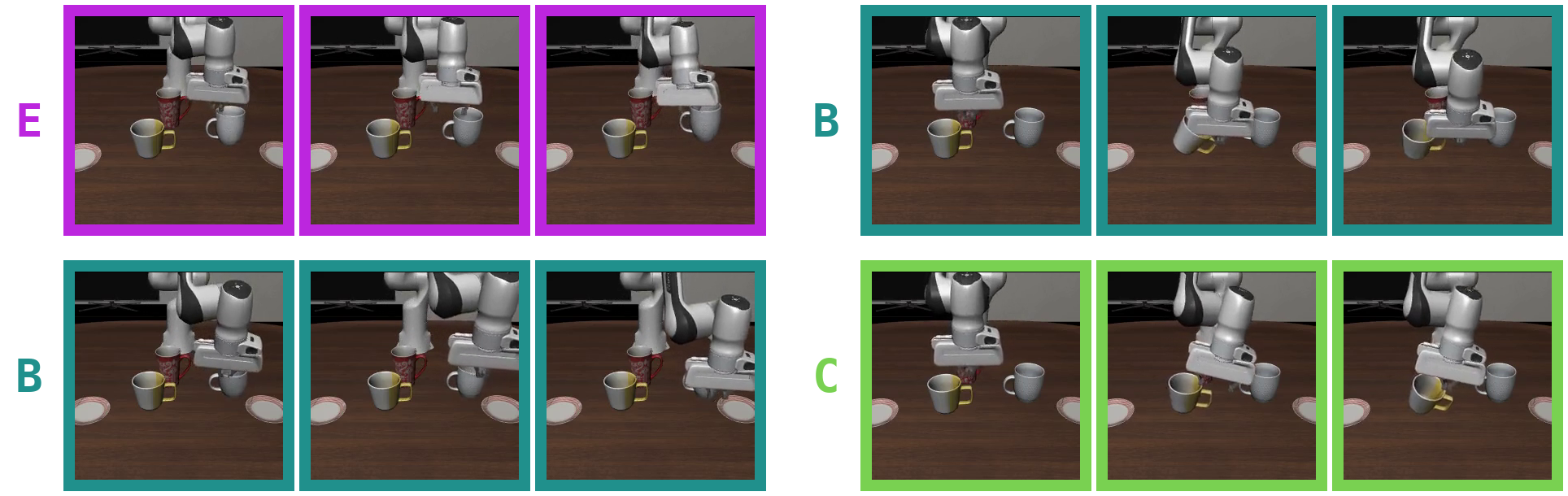}
\caption{\textbf{Manual router substitution for two action chunks.} In both left and right chunks, expert B navigates away from the mug, whereas experts C/E grasp the mug.}    \label{fig:manual}
\end{figure*}
\label{sec:manual_section}

We have shown that the experts selected by the router perform qualitatively different skills that are reused across tasks. Two hypotheses remain to be tested. First, the experts themselves may genuinely differ or they may share overlapping capabilities that the router consistently dispatches in the same way, in which case the apparent one-to-one mapping between experts and skills would be an artifact of the router's policy. Second, even if experts do specialize, their skills may or may not generalize to contexts where the router would not normally select them.

In order to test both of these hypotheses, we substitute in different router assignments for a given chunk within a trajectory. If experts meaningfully differ, we should see qualitatively different behaviors. If an expert corresponds to a reusable skill, we would expect the expected skill behavior even in contexts in which the router would not have selected the expert. These two properties are what we observe empirically. For example, in task 4, Expert C displays grasping behavior even it was never used by the router in this task; we do not observe this behavior when commanding other experts in the same chunk (Figure ~\ref{fig:manual}). As another example, we can recover from a grasping failure where the gripper moved past the object by selecting the grasp expert again, leading to a successful grasp. These results show that the expert primitives are valuable on their own: while our preliminary router failed to recover compositional generalization, manual routing achieved better skill stitching, and suggests out-of-distribution generalization.

\subsection{Performance on LIBERO}
\label{sec:results:libero}

We compare our MoE policy against a dense baseline on two pretrained VLA backbones, $\pi_{0}$ and SmolVLA. In both cases, the MoE and baseline are initialized from the same pretrained checkpoint and finetuned for 20K steps with identical optimization, differing only in whether the action-expert FFN is replaced by the MoE (Sec.~\ref{sec:approach}). 

Our MoE achieves performance comparable to the fine-tuned, dense baseline while, as shown in Sec.~\ref{sec:expert_skills}, learning meaningfully specialized action experts. Thus, a simple MoE architecture drives skill specialization at no cost to task competence.

\section{Conclusion}

In this paper, we show that VLAs trained with a Mixture-of-Experts action expert can learn to decompose manipulation tasks into a small set of modular skills that are reused across different tasks without any pre-defined hierarchy, skill library, or sub-task labels. The router acts as a high-level sequencer, solving longer-horizon tasks by stitching the same primitives in different orders (e.g., grasp-then-place across moka pot, mug, and book tasks), and even failures take the form of repeated application of known primitives rather than degenerate behavior, making the policy interpretable. We learn this compositional structure as a self-supervised, emergent aspect of standard imitation learning with an auxiliary load-balancing penalty, while matching the performance of the VLA baseline. Together, these results suggest that hierarchical, skill-based learning may not require additional machinery (e.g., high-level planners, options, or pre-trained libraries of skills) and instead we can learn such desirable properties from demonstration data alone.

\textbf{Limitations.} Although we have shown qualitative evidence of emergent compositional skill-learning behavior in VLAs, many learned skills still remain task-specific and uninterpretable. Also, experts can sometimes perform behaviors unrelated to their associated primitives, suggesting that the mapping from experts to primitives is not fully precise/discernable. More work is required to understand the essential components of our approach that enable compositionality and how to unlock it even further.

\textbf{Future Work.} We believe this work is an important step in developing data-driven robotic policies that can solve a general set of tasks while also simultaneously learning useful, reusable, atomic skills from data alone. Future work will focus on understanding how these skills can be used to compositionally generalize to longer-horizon tasks and improving upon the architecture to better control what kind of skills are learned.

\newpage
\section*{Impact Statement}
This paper presents work whose goal is to advance the field of machine learning. There are many potential societal consequences of our work, none of which we feel must be specifically highlighted here.

\newpage

\end{document}